\documentclass[conference]{IEEEtran}

\IEEEoverridecommandlockouts

\usepackage{cite}

\usepackage{amsmath,amssymb,amsfonts}

\usepackage{graphicx}

\usepackage{textcomp}

\usepackage{xcolor}

\usepackage{booktabs}

\usepackage{tabularx}

\usepackage{array}

\usepackage{multirow}

\usepackage{url}

\graphicspath{{hidden_emotion_figures/}}

\begin{document}

\title{Beyond Multimodal Emotion Conflict: Temporal Context-Aware Hidden Emotion Detection with Multimodal Large Language Models}

\author{
\IEEEauthorblockN{
Yonghao Wang\textsuperscript{1}
and
Jiaxin Zhou\textsuperscript{2} 
}
\IEEEauthorblockA{\textsuperscript{1}The Chinese University of Hong Kong, China; \textsuperscript{2}Lingnan University, China}
\IEEEauthorblockA{wangyonghao@link.cuhk.edu.hk}
}

\maketitle

\begin{abstract}

Multimodal emotion recognition commonly combines textual, acoustic, and visual signals to infer affective states. Recent work has begun to study conflicts among modalities, but a conflict does not necessarily indicate that a person is concealing an emotion. It may also result from mixed emotions, individual expressive styles, or contextual factors. This proposal aims to distinguish ordinary emotional inconsistency from emotion concealment and to infer the underlying emotion when supported by evidence. We plan to construct a multimodal benchmark with explicit annotations of expressed emotion, internal emotion, and concealment conditions, and to develop an MLLM-based framework that integrates fine-grained facial Action Units (AUs), cross-modal features, and temporal context.

\end{abstract}

\section{Background and Motivation}

The global mental health situation continues to deteriorate, yet many people do not take mental health issues seriously enough. Since the outbreak of the COVID-19 pandemic, societal mental health problems have become even more severe. Despite the increasing demand for mental health services, the number of people actually receiving treatment remains very limited \cite{1}. Therefore, innovative solutions for mental health care are urgently needed.

Currently, there is growing attention on the significant role of AI in the field of mental health. AI mental well-being coaching (AI-MWC) systems, which utilize machine learning algorithms and data analysis, can offer more personalized and efficient mental health support. Such systems need to understand human emotional states and identify users' emotions accurately. Thus, emotion recognition is an important step in affect-aware mental health support. Current research has expanded from single-modal approaches \cite{2,3,4,5,6,7} to multimodal approaches \cite{8,9,10,11,12,13}, achieving significant progress. However, an important challenge remains: people may intentionally or unintentionally mask or regulate their emotions, causing their outward expressions to differ from their internal feelings \cite{14}. For example, someone who feels sad may force a smile, or a person experiencing fear may attempt to maintain a neutral expression. In multimodal settings, such situations may appear as inconsistencies among facial expressions, speech, and textual content. Importantly, however, a discrepancy among modalities does not by itself establish that an emotion is being concealed; it may also arise from mixed emotions, individual expressive styles, or contextual factors. Recent studies have begun to investigate multimodal emotion conflicts in MLLMs. CA-MER examines emotion reasoning when visual and acoustic cues provide conflicting evidence, while MoSEAR seeks to mitigate modality bias and improve balanced modality integration \cite{44}. EmoMM further benchmarks MLLMs under modality-aligned, conflict, and missing-modality settings, and proposes CHASE to steer attention when conflicts are detected \cite{45}.
\begin{figure}[t]\centering

\includegraphics[width=\columnwidth]{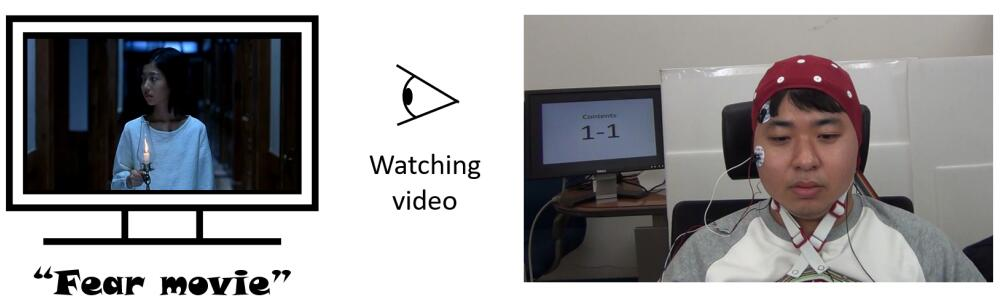}

\caption{A representative example \cite{15} of hidden emotion. A subject watching a horror movie maintains a neutral facial expression but experiences fear internally.}

\end{figure}
These studies make important progress in conflict-aware multimodal emotion recognition, but their primary focus is on handling conflicting or missing modality information for emotion prediction, rather than determining whether a conflict reflects a person's deliberate concealment of an internal emotion. In addition, the existing hidden emotion detection work based on video and electroencephalogram (EEG) signals \cite{15} requires specialized equipment, limiting its practicality for AI-MWC systems. There is also a lack of a dedicated multimodal dataset that jointly represents outward emotional expressions, internal emotional states, and emotion-concealment conditions (Challenge 1). A further challenge is to develop models that can distinguish genuine emotion concealment from ordinary multimodal inconsistency and infer the underlying emotion when sufficient evidence is available (Challenge 2).

Recent advancements in multimodal large language models (MLLMs) provide new opportunities to address Challenge 2. EmoLLMs \cite{4} is a family of open-source instruction-tuned LLMs developed for comprehensive affective analysis. Its multi-task capabilities support the prediction of sentiment strength and emotion intensity, and provide a useful foundation for analyzing nuanced emotional expressions. Meanwhile, conflict-aware studies such as CA-MER \cite{44} and EmoMM \cite{45} show that MLLMs may be biased toward particular modalities when emotional cues disagree. These findings motivate us to go beyond simply selecting or reweighting modalities and investigate whether the observed disagreement is associated with emotion concealment.

This proposal aims to construct a multimodal dataset for hidden emotion recognition, incorporating facial action unit (AU) annotations to represent fine-grained facial dynamics. It will then develop an MLLM-based emotion analysis model that uses information from multiple modalities to distinguish emotion concealment from ordinary emotional inconsistency and, where supported by the evidence, infer the underlying emotion. Building on multimodal affective analysis techniques and the capabilities of MLLMs, the proposed approach will investigate how cross-modal relationships and contextual information can improve hidden emotion recognition in realistic scenarios.

\section{Research Objectives}

The objective of this proposal is to develop MLLM-based sentiment analysis mode for accurate hidden emotion recognition, and apply it to a mental well-being coaching system that can evaluate users' emotional states and provide psychological counseling through interaction, thereby promoting users' mental well-being, which includes the following three objectives:

\begin{itemize}

\item Construct multimodal sentiment datasets that include hidden emotion.

\end{itemize}

\begin{itemize}

\item Design a multimodal feature fusion strategy.

\end{itemize}

\begin{itemize}

\item Validate and evaluate the performance of the optimized MLLM.

\end{itemize}

\begin{figure}[t]\centering

\includegraphics[width=\columnwidth]{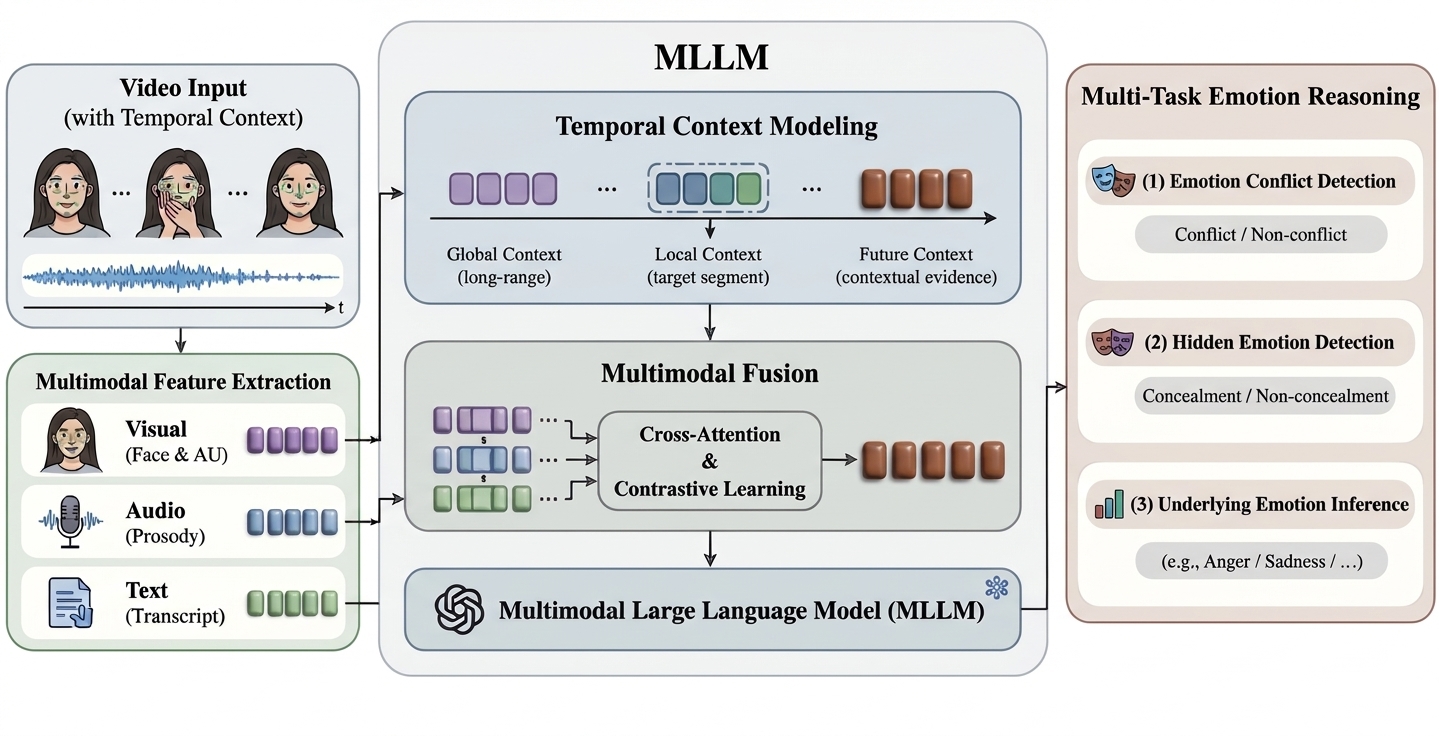}

\caption{Overview of the proposed multimodal hidden emotion detection framework.}

\end{figure}

\section{Literature Review}

\subsection{Multimodal Affective (Sentiment \& Emotion) Analysis}

Research \cite{16} shows that in daily human interactions, emotions are primarily conveyed through facial expressions (55\%), voice (38\%), and text (7\%). Multimodal affective analysis systems \cite{17,18,19,20,21} integrate information from different modalities to provide more accurate emotion analysis for facial emotion recognition (FER) or AU recognition tasks. DFF-ATMF \cite{17} proposed new multi-feature fusion strategies and multi-modal fusion strategies to learn features for text modality and features for audio modality, a multimodal attention fusion module is used. SFNN \cite{18} proposed a neural network based on semantic feature fusion. It uses convolutional neural networks and attention mechanisms to extract visual features, which are mapped to text features and combined with them for sentiment analysis. TIMF \cite{19} learns features separately for each modality and performs tensor fusion of the features. In the decision fusion stage, the upstream results are fused by soft fusion to adjust the decision results. Abbasi et al. \cite{20} proposed an AU-based methodology for child facial expression recognition method specifically that utilizes facial action units rather than raw images or videos and constructs three separate length-independent representations, namely, statistical, spectral and graph at video-level for detailed multi-level facial behaviour decoding. TEDT \cite{21} uses text as the primary information and sound\&image as the secondary information, and proposed a modality reinforcement cross-attention module to convert non-natural language features into natural language features to improve their quality.

Recent studies have specifically examined emotional conflicts among modalities in MLLM-based emotion recognition. Han et al. \cite{44} introduced CA-MER, a benchmark containing video-aligned, audio-aligned, and consistent subsets, and proposed MoSEAR to reduce modality bias through modality-specific experts and attention reallocation. Sun et al. \cite{45} proposed EmoMM, which evaluates MLLMs under aligned, conflicting, and missing-modality conditions. They further identified Video Contribution Collapse (VCC), where MLLMs under-utilize visual evidence, and proposed Conflict-aware Head-level Attention Steering (CHASE) to mitigate this bias. These works demonstrate that modality conflict is an important challenge for reliable multimodal emotion reasoning. Nevertheless, recognizing or resolving a modality conflict is different from recognizing emotion concealment: the same outward inconsistency may occur without deliberate masking, while concealment may not always produce an obvious conflict in every modality.

Although existing works have studied affective analysis across diverse modalities, relatively few have directly addressed hidden emotion recognition. Song and Kim \cite{15} proposed a hidden emotion detection (HED) framework that uses CNNs to analyze video and EEG feature maps. However, EEG acquisition requires specialized equipment and incurs additional implementation costs. More recently, CA-MER \cite{44} and EmoMM \cite{45} have investigated multimodal emotion conflicts in MLLMs, but they focus on modality conflict, modality bias, and missingness rather than explicitly identifying whether the observed discrepancy is caused by emotion concealment. This distinction motivates a benchmark that separately annotates modality inconsistency, outward expression, and the underlying emotional state.

\subsection{Language Model-based Affective Analysis}

In recent years, many studies \cite{22,23,24} have focused on fine-tuning pre-trained language models to enhance their capabilities in the field of affective analysis. Bello et al. \cite{22} combine BERT with other DNNs to improve the ability of the model in short and simple text sentiment analysis. Liao et al. \cite{23} propose a multi-task affective analysis model based on RoBERTa to extract features from both text and aspect tokens, and apply the cross-attention mechanism to guide the model to focus on the features most relevant to the given aspect category. Yin et al. \cite{24} propose the SentiBERT model that integrates a recursive constituency tree based on BERT to better capture compositional sentiment semantics.

Recent studies \cite{3,4,25,26,27,28} have embarked on investigating the utilization of LLMs and resulted in remarkable performance gains in affective analysis tasks. Zhang et al. \cite{25} proposed a retrieval-augmented LLM for financial sentiment analysis, which utilizes additional background information from external sources and outperforms LLM baselines by 15\% and 48\%. They leverage a unique set of instruction-following examples for realigning LLMs to respond more accurately to user-intended financial sentiment analysis tasks. By tuning on multimodal emotional dialogues (i.e., texts and videos), Zhang et al. \cite{27} developed a context and emotion knowledge-tuned LLM named Dialoguellm, which achieved SOTA results on three emotion recognition in conversations (ERC) datasets. Liu et al. \cite{4} proposed the first multi-task affective analysis instruction dataset (AAID) and a comprehensive affective evaluation benchmark (AEB). Based on them, they proposed a open-source LLM family for multi-task affective analysis named EmoLLMs, which is able to predict sentiment strength and emotion intensity and achieves ChatGPT-level and GPT-4-level general capabilities. EmoLLMs possess significant potential to assist us in advancing multimodal data processing and hidden emotion recognition, could bring important technological innovations in the field of affective analysis.

\subsection{Research Gap}

Traditional DNN-based methods \cite{17,18,19,20,21} typically formulate affective analysis (e.g., facial emotion recognition) as classification or regression tasks. Although these methods have achieved progress in multimodal feature extraction and fusion, they may provide coarse-grained emotion descriptions and have limited capacity for contextual reasoning. More importantly, conventional emotion recognition generally predicts the emotion expressed in the available signals, without explicitly modeling the possible difference between an individual's internal emotional state and outward expression.

Recent MLLM-based methods have started to address the problem of inconsistent emotional cues across modalities. CA-MER and MoSEAR \cite{44} benchmark and mitigate emotion conflicts by improving the integration of visual and audio information. EmoMM \cite{45} further studies conflict and missingness and introduces CHASE to reduce modality-related decision bias. These studies establish important benchmarks and methods for robust multimodal emotion reasoning. However, their objective is to improve emotion recognition under conflicting or incomplete inputs; they do not explicitly distinguish ordinary emotional inconsistency from deliberate emotion concealment, nor do they jointly evaluate whether a model can infer the concealed internal emotion. Therefore, the key research gap is not simply the lack of methods for handling modality conflict, but the lack of a clearly defined hidden-emotion recognition task and benchmark that can determine when cross-modal inconsistency is associated with concealment and when it is not. Addressing this gap requires multimodal data with annotations of both outward expressions and independently established internal emotional states, together with a model capable of reasoning over fine-grained and contextual evidence.

\section{Methodology}

This proposal hypothesizes that emotion concealment may lead to discrepancies between internal emotional states and observable cues in one or more modalities. However, such discrepancies are not sufficient evidence of concealment, since they may also result from mixed emotions, individual differences, or contextual factors. Therefore, hidden emotion recognition should be formulated as a task that distinguishes emotion concealment from general multimodal inconsistency, rather than simply treating a conflicting modality as noise. Accordingly, this proposal aims to construct a multimodal dataset with explicit annotations of concealment conditions (Challenge 1), and to design an MLLM-based model that jointly analyzes cross-modal and contextual evidence to detect concealment and infer the underlying emotion when justified (Challenge 2).

\subsection{Dataset Construction}

The quality of training data is one of the most direct factors affecting the model's recognition capability. There are two approaches to constructing a hidden emotion recognition dataset:

\textbf{Direct Construction}: Field collecting \cite{29,30} or extracting from movies or TV episodes \cite{31,32} is a common method for constructing emotion datasets. Considering the inclusion of hidden emotions in the data, it is difficult to filter and annotate data for our objective. Following such initial filtering, human annotators can be hired for fine-grained labeling. Hiring professional actors for filming can yield higher-quality datasets, but it is time-consuming and costly.

Utilizing these tools, we can simulate controlled modality discrepancies in existing emotion datasets (e.g., CMU-MOSEI \cite{35}) for auxiliary robustness experiments, including facial-expression editing and emotion-controlled speech generation techniques \cite{33,34}. Such edited samples can help evaluate how models respond to conflicting cues, but they should not automatically be treated as genuine emotion-concealment instances, because editing a modality does not establish that a person intentionally masked an internal emotion. For the main hidden-emotion benchmark, concealment labels should be supported by the data-collection protocol, participant reports, or other independent annotations.

\subsubsection{Construction of Hidden Emotion}

\begin{figure}[t]\centering

\includegraphics[width=\columnwidth]{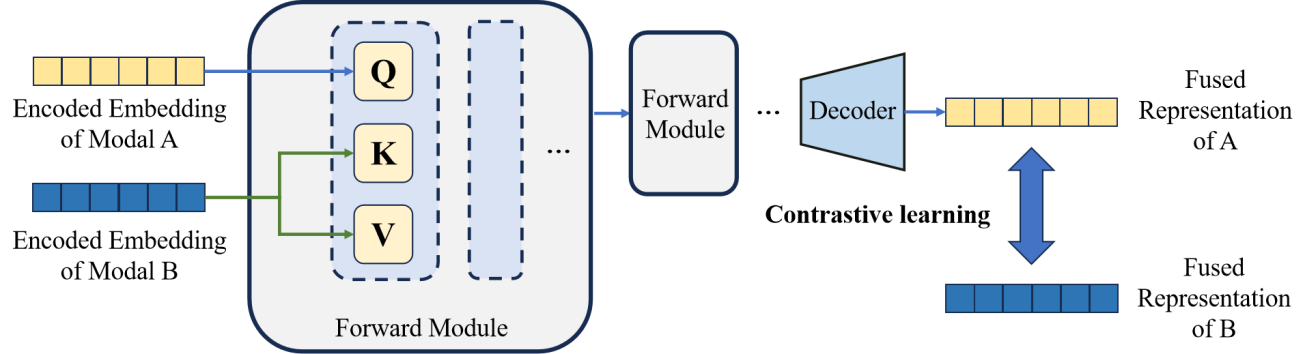}

\caption{Multimodal fusion module with contrastive learning.}

\end{figure}

Using CMU-MOSEI as an example, we define a multimodal data instance as x = \{x\_v, x\_a, x\_t\}, where x\_v, x\_a, and x\_t represent the visual, audio, and textual inputs, respectively. The instance is associated with an expressed-emotion label y\_e and, where available, an independently obtained internal-emotion label y\_i, together with a concealment label y\_c. To construct controlled conflict samples, a specific modality may be modified to convey a different emotion (e.g., changing a sad facial expression to a neutral expression). These edited samples can be used to study modality-conflict robustness. However, they only serve as proxies for cross-modal inconsistency and cannot by themselves establish genuine emotion concealment. For genuine hidden-emotion samples, the internal emotion and concealment condition must be supported by participant reports or the experimental protocol. The model will be trained to predict the relevant labels using a multi-task objective.

\subsubsection{Facial Action Units Annotation}

Compared with raw images or videos, facial action units (AUs) \cite{36} can provide an objective and descriptive representation for measuring facial behavioural changes \cite{37}, which are suitable for our hidden emotion recognition task. We extract AU annotations from videos and uses LLM to describe action units and emotion of corresponding faces, which more information of sentiments for training.

\subsection{Multimodal Fusion Network with Contrastive Learning}

Multimodal user inputs such as audio, text, and expressions provide richer emotional cues. To handle multimodal inputs, this proposal proposes the application of different encoder structures tailored to each modality to effectively extract information. For instance, ResNet networks can be used for encoding video frame inputs, while pretrained LLMs can be used to encode textual information.

After extracting input information from different modalities, the model should be able to filter out noise from certain modalities. To achieve a deeper integration of information from different modalities, this proposal propose to design a Multimodal Fusion Network with cross-attention modules for interaction between the different inputs before they are sent into MLLM as embedded representations.

Referring to the construction of the hidden emotion dataset, it can be assumed that there are emotional differences between the noisy modality and other modalities. Therefore, we plans to utilize contrastive learning to amplify the differences between the representations of noisy modality and others. Consequently, this proposal aims to design the structure illustrated in Fig. 3. Ideally, by fully leveraging the multimodal information, this training approach can enhance the model's robustness against noisy information.

\subsection{Multi-task Emotion Recognition based on MLLMs}

Following the method of EmoLLMs \cite{4}, we can fine-tune the pre-trained LLMs on specially designed multi-task setting through instruction prompts rather than focus only on emotion categorization tasks. It means that in addition to performing multimodal fusion, we also need to annotate the labels of constructed dataset under multi-task settings.

Specifically, the paper designs five sub-tasks for comprehensive emotion analyze: (a) emotion intensity regression (EI-reg), (b) ordinal classification of emotion intensity (EI-oc), (c) sentiment regression (S-reg), (d) ordinal classification of sentiment (S-oc), and (e) emotion classification (E-c). Among them, the E-c sub-task is set to detect multiple emotions contained within the same text, which could provide a strong support for hidden emotion recognition. A detailed instruction prompt template is shown in Table 1.

\begin{table}[t]\centering\caption{Templates for constructing prompts for the instruction dataset.}

\begin{tabular}{lp{0.68\columnwidth}}\toprule

\textbf{Task} & \textbf{Prompt Template}\\\midrule

EI-reg & Task: [task prompt] Input: [fused representation] Emotion E: [emotion] Intensity score: [output]\\

EI-oc & Task: [task prompt] Input: [fused representation] Emotion E: [emotion] Intensity class: [output]\\

S-reg & Task: [task prompt] Input: [fused representation] Intensity score: [output]\\

S-oc & Task: [task prompt] Input: [fused representation] Intensity class: [output]\\

E-c & Task: [task prompt] Input: [fused representation] This input content contains emotions: [output]\\

\bottomrule\end{tabular}\end{table}

\begin{figure}[t]\centering

\includegraphics[width=\columnwidth]{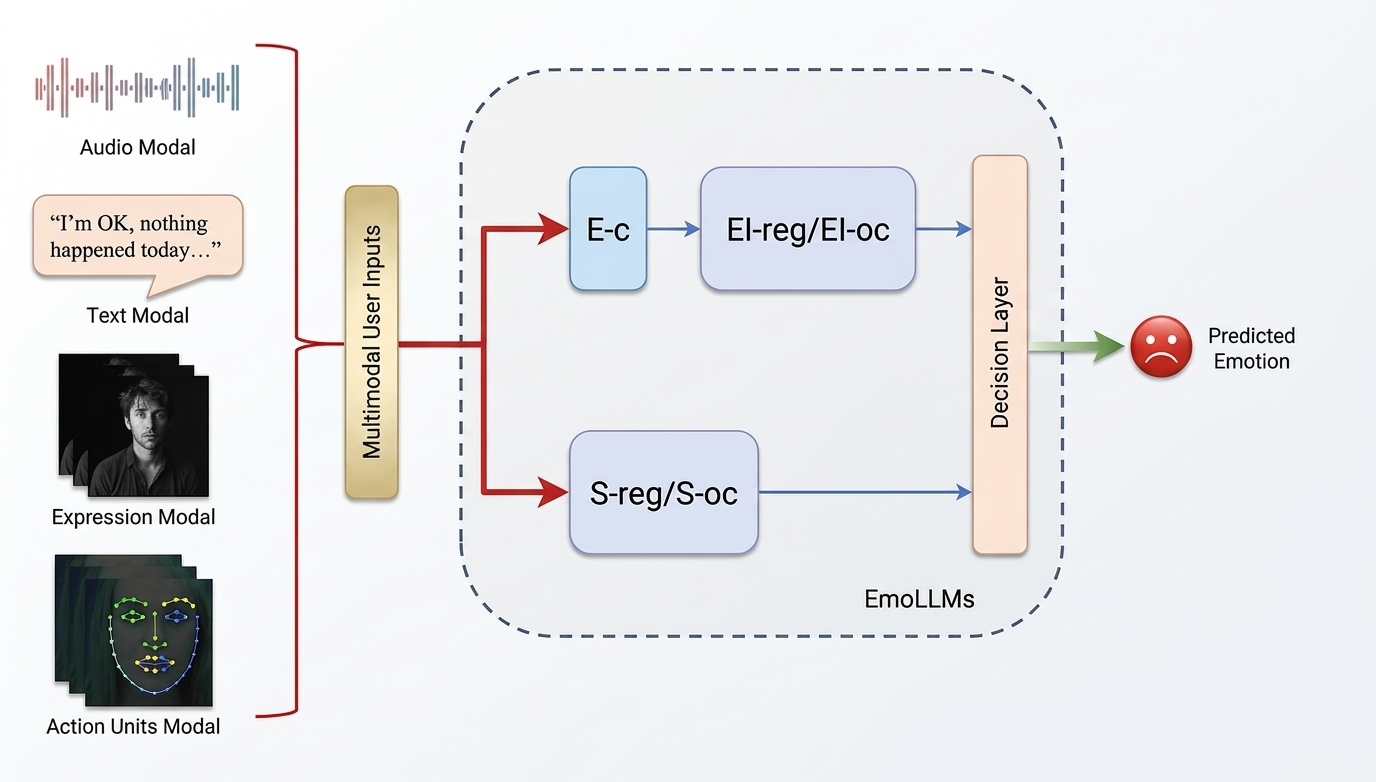}

\caption{Combining EmoLLMs sub-tasks for hidden emotion recognition.}

\end{figure}

Therefore, the combination of these sub-task functions provides a potential possibility for hidden emotion recognition. As shown in Figure 4, the model firstly analyses the multiple emotions contained in the input information (E-c), and then predicts the intensity of each emotion individually analyzed by E-c (EI-reg/EI-oc). Finally, combine the predicted sentiment intensities defined in S-reg/S-oc task (positive or negative), the model is expected to identify the final hidden emotion and give a comprehensive analysis conclusions.

\subsection{Model Training}

We can conduct our model training in two stages:

Multimodal Fusion Network Pre-training: this proposal propose to pre-train the Multimodal Fusion Network as a pre-processing of constructed dataset before official MLLM tuning. The trained network is used to provide appropriate fused representations as the inputs, and can be seen as the input embedding encoder module of the MLLM.

MLLM Fine-tuning: Given fused embedded representations, we fine-tune some specified the pre-trained MLLMs on our designed multi-task setting to predict sentiment intensity or emotion category. Some commonly used training strategies (e.g. self-supervised learning \cite{38}) can be used for our fine-tuning.

\subsection{Model Evaluation and Performance Validation}

\subsubsection{Model Validation Data Sets}

Standard Emotion Recognition: evaluation will be performed on several popular and publicly available multimodal sentiment datasets, such as the MuSe-CaR dataset \cite{39}, the Representative MOSI dataset \cite{40} and the MELD dataset \cite{41}. And a comprehensive Affective Evaluation Benchmark (AEB) \cite{4} containing 14 affective tasks from different sources and different domains will be used to validate the generalization ability of the optimized model, and the performance of the optimized model will also be compared with other existing methods.

Hidden Emotion Recognition: the most straightforward approach is to use cross-validation to verify the performance of our model, or attempt to apply the method described in Section IV (A) to modify other related datasets. This allows for validation of the model on a broader range of data and examination of its generalization capability.

\subsubsection{Model Validation Metrics}

This proposal will use the official evaluation metric \cite{42} as the evaluation metric for EI-reg, EI-oc, S-reg, and S-oc and use accuracy, micro-F1 (mi-F1), macro-F1 (maF1) for E-c. These metrics will be combined to validate the performance of the model in hidden emotion recognition.

\subsection{A little idea for the future}
Hidden emotion recognition should not rely solely on static facial expressions or isolated behavioral cues, as emotional concealment may manifest through subtle changes over time. By analyzing longer video sequences and incorporating temporal context, including emotional expressions and behavioral variations before, during, and after a target interaction, it may be possible to better identify inconsistencies between outward expressions and underlying emotional states. Such long-range temporal modeling could provide richer contextual evidence for hidden emotion inference, while continual learning \cite{43} may further help capture individual-specific emotional expression patterns over time.

\bibliographystyle{IEEEtran}

\bibliography{reference}

\end{document}